# TOKON: TOKenization-Optimized Normalization for time series analysis with a large language model

Janghoon Yang
*Computer Science Program.*
*Penn State Abington*
Abington, PA, USA
jxy5427@psu.edu

*Abstract*— While large language models have rapidly evolved towards general artificial intelligence, their versatility in analyzing time series data remains limited. To address this limitation, we propose a novel normalization technique that considers the inherent nature of tokenization. The proposed Tokenization-Optimized Normalization (TOKON) simplifies time series data by representing each element with a single token, effectively reducing the number of tokens by 2 to 3 times. Additionally, we introduce a novel prompt for time series forecasting, termed Time Series Forecasting with Care (TFSC), to further enhance forecasting performance. Experimental results demonstrate that TOKON improves root mean square error (RMSE) for multi-step forecasting by approximately 7% to 18%, depending on the dataset and prompting method. Furthermore, TFSC, when used in conjunction with TOKON, shows additional improvements in forecasting accuracy for certain datasets

*Keywords— Normalization, prompting, large language model, time series, forecasting, prediction*

## I. INTRODUCTION

With the evolution of deep learning in natural language processing, the ubiquitous nature of large language models (LLMs) is becoming increasingly robust [1]. Initially introduced for Q&A services, these models can now be applied to sound and image modalities, enabling LLMs to understand and integrate multi-modal information [2]. The extensive scale of these models allows them to achieve state-of-the-art (SOTA) performance across various tasks, including natural language understanding, event recognition, and coding. However, publicly available LLMs often exhibit limited performance in handling time series data, despite recent research efforts to address this limitation [3].

To enhance LLM versatility for time series tasks, approaches such as direct prompting, fine-tuning, and tokenization have been explored. Methods include converting time series to text for anomaly detection [4], using LST prompts to decompose forecasting into short-term and long-term components [5], and integrating external knowledge into prompts [6]. Fine-tuning LLMs involves training with question-answer pairs for forecasting [7] and optimizing parameters in two stages: fitting to time series data and further fine-tuning with parameter-efficient fine-tuning (PEFT) for specific downstream tasks [8]. Time-series-specific tokenization methods have also been studied, including decomposing time series into components and applying patching separately [9], and reprogramming LLMs by converting time series into text prototypes and augmenting the patches with it [10]. However, direct prompting often fails to achieve SOTA performance, while fine-tuning and tokenization can be excessively complex, limiting their use by researchers with constrained computing resources.

Normalization is another crucial aspect of time series analysis. When models are trained on datasets from different domains, normalization can enhance numerical stability, especially in gradient-based optimization algorithms. Deep learning models benefit from better convergence by avoiding saturation with nonlinear activation functions and improved generalization by mitigating domain-specific data ranges. SOTA models for time series tasks leverage normalization to enhance performance. For instance, normalizing input time series to unit norm improves forecasting and classification performance [11]. Learnable normalization techniques, such as hybrid normalization modules for anomaly detection [12] and adaptive normalization layers for prediction [13], significantly boost performance. Various normalization methods for time series are reviewed in [14]. While some approaches [10] apply normalization before patching, the impact of normalization in patching for LLM time series tasks has not been thoroughly studied yet.

Performance improvements in LLMs through articulated prompting are often limited, and fine-tuning requires substantial computational resources. Time-series-specific tokenization demands significant effort and datasets. Patching, while an alternative, necessitates additional models for adaptation and extensive learning to work with existing LLMs. A less explored direction to enhance LLM performance is leveraging normalization. Conventional normalizations, such as standard or min-max normalization, are limited by the nature of LLM tokenization. This research aims to develop a method for improving time series forecasting performance with LLMs without fine-tuning or additional networks. We propose a normalization method tailored to the tokenization used by LLMs, termed TOKenization-Optimized Normalization (TOKON). Specifically, we utilize GPT-4o-mini by OpenAI, though TOKON is applicable to other LLMs for time series forecasting. Additionally, we introduce a novel prompt for time series forecasting, Time Series Forecasting with Care (TSFC), to further enhance performance by considering LLM behavior during forecasting tasks. Our contributions are as follows:

1. We propose a novel normalization method that considers the tokenizer's nature in LLMs, significantly improving forecasting performance by converting each number to a valid token in the existing tokenizer dictionary. This normalization does not require fine-tuning, additional modules, or tokenizer modifications.

2. The normalization parameter, determined through a 1D search, provides superior performance and is applicable regardless of LLM type.

3. Extensive simulation results demonstrate that TOKON universally improves performance, irrespective of the prompting method.

4. The proposed TFSC prompt further enhances performance when applied with the proposed normalization.

This paper is organized as follows. Section 2 reviews normalization and tokenization of time series for forecasting with LLMs. Section 3 presents the proposed normalization method, TOKON, and the 1D search for determining normalization parameters, followed by an explanation of the TFSC prompt for time series forecasting with LLMs. Section 4 details the experimental setups, including datasets, LLMs, tokenizers, and prompting methods, and demonstrates the efficacy of the proposed normalization and prompting for forecasting tasks. Section 5 concludes with remarks and future directions.

## II. Related Works

### A. Normalization of time series for forecasting with LLM

In machine learning and deep learning, normalizing numeric time series is crucial for training models with generalization capability. Proper normalization can also improve convergence speed by avoiding excessive gradients. Similarly, the performance of time series analysis with large language models (LLMs) is often influenced by normalization. However, unlike deep learning models dedicated to time series tasks, the effect of normalization can vary significantly depending on the representation of the time series, fine-tuning, and additional modules for specific time series tasks.

When time series data is part of a query, it is often unnormalized to preserve information that the LLM can exploit using its world knowledge [7]. However, the time series can be converted into a simpler form to simplify the problem. For example, after deriving percentage changes from the time series, these changes can be quantized into several levels of up and down movements to make the prediction problem easier [15]. Instance normalization is often used when time series data is embedded and aligned to the LLM's space [8][9][16], making the time series task more robust to distribution shifts. While normalization parameters, such as scale and offset, are often learnable and trained [9], it is also common to normalize time series data to have a mean of zero and unit standard deviation without learnable parameters [8][10][16].

### B. Representation of time series

The representation of input to a model is critically significant in enabling the model to exploit important information for task completion. Existing large language models (LLMs) are trained on vast corpora of textual data, allowing them to efficiently represent textual information. Typically, these models tokenize text into units of words or subwords and then represent each token with a high-dimensional feature vector, known as token embedding. Tokenization begins with parsing textual input based on a dictionary. However, this dictionary does not include every possible number as a word, fundamentally limiting the efficient representation of numeric time series. To address this limitation, patching and alignment are often used to represent time series in a way that LLMs can understand. However, the representation of time series data depends on the LLM's structure, task-specific additional modules, and fine-tuning. The formation of input to the LLM with time series embedding also varies depending on the model architecture.

One of the simplest forms of time series representation is to use random mapping of time series to vocabularies in the tokenizer. Random mapping of any data type has been shown to generate valid outputs by leveraging in-context learning in LLMs without requiring additional modules or fine-tuning [17]. Another straightforward method to fine-tune LLMs for time series tasks is to use time series data solely as input without leveraging context information [9]. This type of LLM can be fine-tuned for specific tasks using an adapter module that transforms the patch or time series into the LLM's vector space. Linear layers [9] or convolution layers [8] are often used as adapter modules.

Several different structures can be found when LLMs leverage both textual and time series embeddings. An embedding module can be trained to convert patches of time series data using self-supervised learning, followed by end-to-end fine-tuning where the input is a concatenation of textual embeddings and a sequence of patch embeddings [15]. Similarly, an entire time series can be embedded using a convolutional neural network and concatenated with LLM output for textual information, which is then further processed with a projector for classification tasks [18]. The LLM output with concatenated textual and time series embeddings may also be processed further with regression.

Time series embeddings can be further contextualized by representing them with prototype texts. After embedding the patch with a linear layer, the embedded vector is reprogrammed with prototype texts through a multihead cross-attention layer, aligning the patch embedding with stereotyped textual descriptions. This approach can potentially enable LLMs to better leverage information in time series data [10]. Similarly, contrastive learning over both instance-wise and column-wise dimensions can prevent the representation space from shrinking by augmenting positive and negative samples from each segment of the time series. Intentionally placing embeddings near typical text descriptions of the time series, such as frequency and shape, can help align time series embeddings with preselected prototype texts [19].

## III. Methods

### A. Tokenization-optimized normalization

The LLM's capability to complete tasks through in-context learning with the random mapping of arbitrary data types to tokens [17] demonstrates its potential for time-series forecasting. The near state-of-the-art performance of LLMs fine-tuned using a question-and-answer approach also highlights their capabilities as zero-shot learners [7]. Considering this capability, the zero-shot learning performance of LLMs can be further improved by representing time series data in the prompt in a way that is more compatible with the tokenizer. This can be seen as articulated mapping to tokens rather than random mapping, to better leverage the in-context learning of the LLM.

For example, a series of three floating-point numbers, '1023.37, 950.2, 1111.11,' can be split into tokens using character-level tokenization as ['102', '3', '.', '37', ',', '950', '.', '2', '111', '1', '.', '11']. Forecasting the next number after tokenization introduces several challenges. First, the varying number of tokens per floating-point number can create ambiguity for the LLM, as the model must infer the boundaries and relationships between tokens to reconstruct the original numbers. Second, the effective sequence length increases significantly, making the forecasting task more computationally expensive and challenging, especially for long time series. Lastly, tokenization reformulates the original single-step prediction task into a multi-step prediction problem, where the model must predict multiple tokens in sequence to generate a single number.

To address these issues, we consider normalizing time series data to integers within the dictionary of the tokenizer. For simplicity, we assume that the tokenizer's dictionary contains continuous integers from $I_{min}$ to $I_{max}$. Let the $j$th element at the $i$th series $s_i$ be denoted as $s_{i,j}$. TOKON applied to $s_{i,j}$ can be expressed as:

$$v_{i,j} = \max(\min(r(\sigma_T \frac{s_{i,j} - m_s}{\sigma_s} + m_T), I_{max}), I_{min}) \quad (1)$$

where $m_s$ and $\sigma_s$ are the sample mean and sample standard deviation over all elements for time series in the same domain respectively. $m_T$ and $\sigma_T$ are the target mean and the target standard deviation respectively. The functions $\max(\cdot)$, $\min(\cdot)$, and $r(\cdot)$ represent the maximum, minimum and rounding to nearest integer operations, respectively.

TOKON maps a number in the series to an integer within the tokenizer's dictionary while preserving its ordinal position. The underlying assumption is that an LLM, functioning as a general pattern-matching mechanism, can achieve improved learning from its own prompt when the time series is represented in a more simplified form. With this normalization, each number in a series is represented by a single token, eliminating the need to infer relationships between tokens to reconstruct the original numbers. Additionally, the effective sequence length after tokenization is decreased by typically 2 or 3 times while maintaining the original forecasting task. Thus, LLMs are likely to benefit from this simplified reformulated problem using TOKON.

*B. Parameterization with 1D search*

TOKON requires the specification of two parameters, , $m_T$ and $\sigma_T$. Let the LLM be denoted by $f(x)$ where $x$ is the query to the LLM. We can also denote the cost function by $C(f(x_i), y_{i,T})$ where $y_{i,T}$ is the ground truth to the input $x_i$. The optimal parameters of TOKON will be those that minimize the average cost function. However, $f(x)$ is often a non-convex function, making it very difficult to minimize optimally. To simplify the parameter search, we assume that numeric values are symmetrically distributed. With this assumption, we can safely fix $m_T$ as $(I_{min} + I_{max})/2$.

Focusing on setting $\sigma_T$, a 1D search is considered to find an appropriate parameter. While a 1D search does not guarantee optimal parameterization, it can ensure local optimality even when the problem is non-convex. To implement the 1D search efficiently, the Golden Section Search method, as shown in Figure 1, can be utilized. The corresponding algorithm is depicted in Figure 1. Initially, the range for the 1D search is set from $I_{min}$ from $I_{max}$. Parameters $m_s$, $\sigma_s$, and $m_T$ required for scaling are also calculated. At each iteration, the search interval for optimizing the parameter $\sigma_T$ is updated using the Golden Section Search rule. It determines two probing points, $\delta_1$ and $\delta_2$, which are set as temporary target scaling factors $\sigma_T$. The time series is normalized using parameters $\delta_p$ and $m_T$. Then, it is embedded into a query to create an input for the LLM. To collect responses for each query, the sum costs, $C_{\delta_1}$ and $C_{\delta_2}$, are calculated. Depending on the magnitude of the sum costs, the search space interval is adjusted. Finally, when the interval length is less than or equal to $\varepsilon$, $\sigma_T$ is determined as $(\delta_{max} + \delta_{min})/2$.

While this 1D search is guaranteed to converge, there are some issues that need to be addressed in future research. First, the LLM may not fully exploit its knowledge of physical phenomena. While the relationship between consecutive numbers in a series can be tracked, the LLM may have limited use of world knowledge due to frequently occurring mismatched numbers in some physical phenomena. In extreme cases, its knowledge of physical phenomena can interfere rather than provide helpful information. To avoid this problem, one might try to remove some context information. Additionally, depending on the scale and offset parameters, there can be a tradeoff between quantization error and the complexity of time series pattern.

*C. TSFC Prompt*

Despite efforts to enhance forecasting with articulated prompting, no single powerful prompt works universally across various datasets. The response characteristics of LLMs associated with time series forecasting were presented in [20]. LLMs may struggle to properly manipulate multiple components when asked to decompose a time series into its underlying components. It was also observed that as the number of operations increases, LLMs often fail to perform basic algebraic operations, such as addition and subtraction, accurately. They frequently fail to exploit long-term characteristics like trends and seasonality, relying only on recent values even when definite seasonality is present, despite prompts specifically requesting consideration of these characteristics.

As an alternative approach to address these issues directly in the prompt, a new prompt is designed to request LLMs to analyze time series with consideration of trends and seasonality, rather than decomposing them into components, and to execute basic algebraic operations carefully. This approach is called the Time Series Forecasting with Care (TSFC) prompt. The TSFC prompt can be given as follows: "Analyze the time series step by step, focusing on identifying and leveraging trends and seasonal patterns. Execute each algebraic operation carefully, ensuring precision and accuracy at every stage. Pay close attention to trends and seasonal patterns, especially when determining the final answer". The main design paradigm emphasizes performing the analysis of time series step by step, with a focus on trends, seasonality, and precision.

```
1. Set initial range [δ_min, δ_max] such that δ_min = I_min and δ_max = I_max
2. Calculate m_s and σ_s, and set m_T as (δ_min + δ_max)/2
3. Set golden ratio conjugate, ρ = (√5 − 1)/2
4. δ_1 = δ_min + (δ_max − δ_min)ρ, δ_2 = δ_max − (δ_max − δ_min)ρ
5. for p ∈ {1,2}
     for i ∈ S_1D and p ∈ {1,2}
         Normalize the time series
         v_{i,j}(δ_p) = max(min(r(δ_p (s_{i,j.} − m_s)/σ_s + m_T), I_max), I_min)
         Embed the normalized time series v_{i,j}(δ_p) into the query x_i
         C(f(x_i), y_{i,T})
     C_{δ_p} = Σ_{i∈S_1D} C(f(x_i), y_{i,T})
6. if C_{δ_1} < C_{δ_2}
       δ_max = δ_2
   else
       δ_min = δ_1
7. if δ_max − δ_min > ε, Go to step 4, else, σ_T = (δ_max + δ_min)/2
```

Fig. 1. 1D Search for TOKON parameterization.

## IV. EXPERIMENTS

**Task**: A multi-steps forecasting for a univariate time series will be considered.

**Datasets**: The Average IHEPC (AIHEPC) dataset and a subset of the M4 (SM4) training dataset were used to evaluate time series forecasting performance. The AIHEPC dataset, constructed by averaging the global intensity in the Individual Household Electric Power Consumption (IHEPC) dataset from the UCI Machine Learning Repository over 60 minutes, consists of 3000 series, each with a length of 96. Since it represents hourly current usage, it inherently exhibits some degree of seasonality. The M4 training dataset, with monthly granularity, contains diverse time series representing various economic indicators, such as GDP growth rate and unemployment rate. Although the specific sources of each time series are not publicly disclosed, the economic nature of the data suggests the presence of seasonality, trends, and abrupt changes due to policy shifts. The lengths of time series in this dataset vary significantly. Considering cost, the number of series, and interpretability, a dataset was constructed with 965 series of length 64 and 1104 series of length 49, where each time series is embedded into the prompt with context information.

**Models and Tokenizers:** Considering performance and cost, GPT-4o-mini was selected as the target LLM. This model tokenizes sentences using the Tiktoken tokenizer, which is based on byte pair encoding (BPE). The Tiktokenizer's dictionary includes integers from 0 to 999.

**Promptings:** A baseline prompt was created to include essential context information such as date and time, and the number of elements in the time series. An example of the data in this dataset is: "Given the recorded measurements from 2009-12-01 to 2014-01-01 spanning 49 months, with the values: 1000.0, 1032.0, …., 1197.0, predict the next 18 measurements." To focus on the effect of normalization, the baseline prompt, chain of thought (CoT) prompt, and TSFC prompt were considered instead of experimenting with various few-shot prompting examples. CoT and TSFC prompts were created by concatenating the baseline prompt with CoT-specific and TSFC-specific fixed prompts, respectively. Finally, every prompt is concatenated with "please answer the predicted values only" to simplify parsing the generated output.

**Parameterization with 1D Search:** The scale parameter for normalization was determined at the dataset level. It can be determined at different levels, such as an example level or a corpus level across multiple datasets, which requires further attention in future research. To this end, the first 100 samples were selected from the dataset. From this subset, $m_s$ and $σ_s$ were calculated and used for normalization of other samples in the dataset, while $m_T$ was set as 499.5. As expected, due to the non-convex nature of the problem, the root mean squared error (RMSE) does not decrease monotonically. It was also found that the best performance is not achieved at convergence. Rather than setting a scale parameter as the one achieving the best performance during the iteration, it was set as the average of the lower end and upper end at convergence, following the algorithm description in Figure 1. Consequently, $m_s$ and $σ_s$, and $σ_T$ were set as 4.98, 4.99, and 24.57 for the AIHEPC dataset, and 3724.92, 3145.08, and 312.31 for the SM4 dataset.

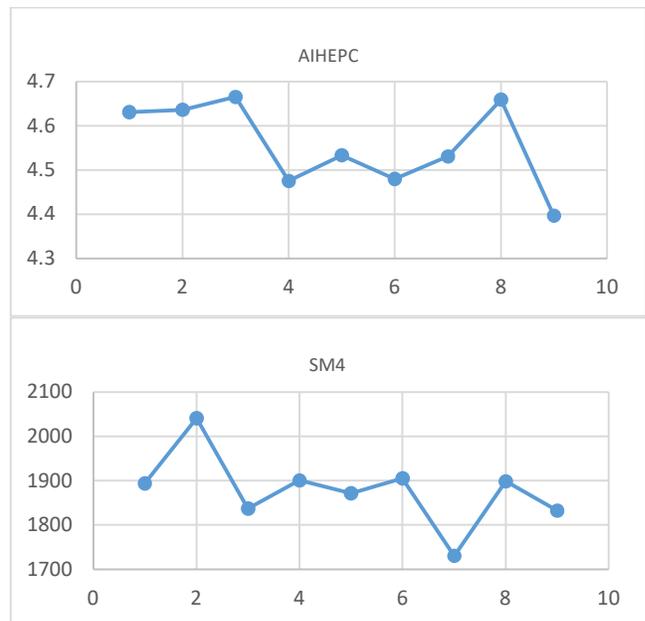

Fig. 2. Convergence characteristic with 1D search to determine scale parameter where x axis is the number of iterations and y axis the RMSD Search for TOKON parameterization.

**Main Results:** To assess the efficacy of the TOKON normalization and TSFC prompting, the proposed methods were compared for different datasets. The performance with the AIHEPC dataset is shown in Table 1. For this dataset, the task is to predict the next 6 steps, with each value in the table representing the average performance over these 6 steps. TOKON improves RMSE performance for baseline, CoT, and TSFC prompting by 7.74%, 13.90%, and 18.60%, respectively, while it improves MAE performance by 8.89%, 10.06%, and 13.36%, respectively. TOKON is observed to enhance performance regardless of the prompting and metrics used. The proposed TSFC prompting also contributes to performance improvement when combined with TOKON. It improves RMSE performance for baseline and CoT prompting by 12.3% and 4.62%, respectively, while it improves MAE performance by 1.70% and 2.29%, respectively. The more significant improvement in RMSE performance suggests that TSFC can effectively reduce large errors by following the directions provided in the prompt.

The performance of the TOKON normalization and TSFC prompting for the SM4 dataset is shown in Table 2. For this dataset, the task is to predict the next 18 steps, with each value in the table representing the average performance over these 18 steps. TOKON improves RMSE performance for baseline, CoT, and TSFC prompting by 19.13%, 13.17%, and 27.83%, respectively, while it improves MAE performance by 16.90%, 12.29%, and 16.82%, respectively. However, the proposed TSFC prompting performs worse when TOKON is not applied, while all considered promptings with TOKON achieve similar performance. This dependency on the dataset can arise from various factors. Different datasets have distinct characteristics in time series, where forecasting with a given prompting may leverage advantageous characteristics, which can be considered as data specificity beyond the task type. Domain-specific knowledge may also affect performance. Even with normalization, if the LLM can exploit domain-specific time series patterns in accordance with the given prompt, it can improve performance. Another potential cause is the forecast horizon. Some prompts may be better suited for short-term forecasting, while others for long-term forecasting. To this end, the performance with the SM4 dataset was compared by averaging over the first 6 steps only in Table-3. While the absolute error reduces due to the high correlation with the last few elements in the series, the gain with TSFC is still found to be limited.

To identify the detailed characteristics of the TOKON and TSFC prompting, the normalized RMSE with TOKON for each forecasting step was plotted in Figure-3. Each RMSE was normalized by the minimum over prompting methods and forecasting steps. For both datasets, the minimum RMSE occurs at the first step. This characteristic is reasonable, as GPT-4o-mini often considers a few recent values for forecasting and the value at the first step is likely to be most correlated with the recent values. For the AIHEPC dataset, the TSFC prompting achieves relatively consistent performance, while other prompting methods show stronger dependency on forecasting steps. This result may be attributed to the effect of the TSFC prompt, which may lead to minimizing variance across all forecasting steps, whereas other prompting methods tend to prefer recency in forecasting. For the SM4 dataset, TSFC prompting shows the best performance at many intermediate steps. Notably, TSFC prompting has the smallest worst performance over forecasting steps. This characteristic may be attributed to its cue from the TSFC prompting to enforce long-term information extensively, which can potentially contribute to forecasting errors arising from not using long-term information properly. Overall, step-wise forecasting results highlight the potential of the proposed TSFC prompting for the forecasting task.

TABLE I. Performance with AIHEPC dataset

|  | RMSE |  | MAE |  |
|---|---|---|---|---|
| Normalization | No | Yes | No | Yes |
| Baseline | 4.621 | **4.356** | 3.472 | **3.163** |
| CoT | 4.651 | **4.005** | 3.539 | **3.182** |
| TSFC | 4.692 | **3.820** | 3.588 | **3.109** |

TABLE II. Performance with Performance with SM4 dataset

|  | RMSE |  | MAE |  |
|---|---|---|---|---|
| Normalization | No | Yes | No | Yes |
| Baseline | 2704.62 | **2187.18** | 1515.36 | **1259.18** |
| CoT | 2545.87 | **2210.49** | 1436.77 | **1260.05** |
| TSFC | 2996.53 | **2162.30** | 1518.52 | **1262.96** |

TABLE III. . Performance with SM4 dataset for the first 6 steps

|  | RMSE |  | MAE |  |
|---|---|---|---|---|
| Normalization | No | Yes | No | Yes |
| Baseline | 2408.38 | **2065.10** | 1312.90 | **1129.28** |
| CoT | 2256.42 | **2099.23** | 1203.86 | **1137.08** |
| TSFC | 2651.51 | **2028.58** | 1306.30 | **1122.24** |

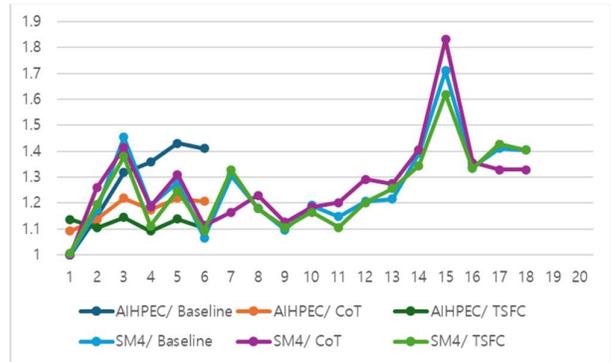

Fig. 3. Normalized RMSE performance with TOKON at each forecasting step for each prompting and each dataset, where RMSE for each dataset is normalized by the minimum RMSE over steps and prompting methods (x-axis represents the steps in forecasting and y-axis the normalized RMSE).

V. CONCLUSIONS

In this paper, a tokenization-optimized normalization and a prompting method for time series forecasting were proposed. The proposed TOKON was found to improve forecasting performance across all considered prompting methods and datasets. TSFC also achieved the best performance for one dataset and near-best performance for the other. The improvement by TOKON is conjectured to result from reducing the number of tokens in time series by 2 to 3 times, representing each element in the time series with a single token, which simplifies in-context learning. The lowest worst-case forecasting performance with TSFC prompting for both datasets also shows its potential as a promising prompt for forecasting.

However, several issues need to be addressed in future research. The proposed TSFC needs to be tested on diverse datasets, and its performance can be further improved with articulated few-shot examples. TOKON needs to be tested on different tasks, such as classification and outlier detection, to assess its efficacy across various time series tasks. TOKON may produce large errors when forecasting outlier values in a time series. A prompting method can be developed to refine the final forecasting using domain-specific knowledge beyond the range limited by the numbers in the dictionary. To further boost performance with TOKON, fine-tuning can be considered to enhance in-context learning capability in time series analysis. More importantly, technical issues in multivariate time series, such as presenting multivariate time series as queries and scaling considering heterogeneity in scale across different variables, need further attention.